\documentclass[a4paper,11pt]{article}

\usepackage[psamsfonts]{amssymb}
\usepackage{latexsym}
\usepackage{setspace}
\usepackage[pdftex]{graphicx}
\usepackage{subfig}
\usepackage{amsmath}

\title{Variable subset selection via GA and information complexity in mixtures of Poisson and negative binomial regression models}

\author{ \centerline{ \bf\small TYLER J. MASSARO}\\  
\centerline{\small\it Department of Mathematics}  \\
\centerline{\small\it University of Tennessee, Knoxville, 37996-0532, TN, USA} \\
\centerline{\small\it E-mail: massaro@math.utk.edu} \\
~\\
\centerline{ \bf\small HAMPARSUM BOZDOGAN}\\  
\centerline{\small\it Department of Business Analytics \& Statistics}  \\
\centerline{\small\it University of Tennessee, Knoxville, 37996-0532, TN, USA} \\
\centerline{\small\it E-mail: bozdogan@utk.edu}
}
\date{}

\begin{document}

\maketitle

\begin{abstract}
Count data, for example the number of observed cases of a disease in a city, often arise in the fields of healthcare analytics and epidemiology.
In this paper, we consider performing regression on multivariate data in which our outcome is a count.
Specifically, we derive log-likelihood functions for finite mixtures of regression models involving counts that come from a Poisson distribution, as well as a negative binomial distribution when the counts are significantly overdispersed.
Within our proposed modeling framework, we carry out optimal component selection using the information criteria scores AIC, BIC, CAIC, and ICOMP.
We demonstrate applications of our approach on simulated data, as well as on a real data set of HIV cases in Tennessee counties from the year 2010.
Finally, using a genetic algorithm within our framework, we perform variable subset selection to determine the covariates that are most responsible for categorizing Tennessee counties.
This leads to some interesting insights into the traits of counties that have high HIV counts.
\end{abstract}


\section{Introduction}
Count data often arise in healthcare and epidemiology data sets.
For example, the number of outcomes observed (\emph{i.e.} cases of a disease) in a specific group of people (\emph{i.e.} citizens residing in Knox county).
We tend to treat count data as realizations from a Poisson distribution, so that the probability of observing $y$ events given an expected number of events, $\mu$, is precisely
\begin{equation}
\mathbb{P}[Y = y; \mu] = \frac{\mu^y \exp(-\mu)}{y!}.
\end{equation}

In regression analysis, Poissonian data are most often related to a linear combination of independent variables (\emph{i.e.}, covariates, exposures) via a log-link function.
Hence, the Poisson regression model is defined as
\begin{equation} \label{poissmodel}
\ln\left(\mathbb{E}[Y]\right) = \beta_0 + \sum_{j=1}^p \beta_j X_j,
\end{equation}
where $\beta_j$, $j = 1, \dots, k$, are the measured effects of $k$ indpendent exposures, $X$.
For a binary exposure, $X_j$, the amount $\exp(\beta_j)$ is the increase (or decrease) in outcome incidence for subjects in which the exposure was observed, relative to subjects in which the exposure was not observed \cite{Dohoo:2012}.
The amount $\exp(\beta_0)$ is the expected number of counts in baseline subjects.
Researchers are responsible for determining the criteria defining a baseline measurement.

A major assumption in Poisson regression modeling is that the mean and variance of the observed counts are equivalent.
When this assumption is violated, as is often true of real data, we say the data are overdispersed.
A negative binomial distribution may more appropriate than a Poisson for describing overdispersed count data, due to its excessive tail behavior \cite{Dohoo:2012}.

The difference between negative binomial regression (also referred to as NB-2 models) and Poisson regression comes from our treatment of the observed count data.
We still make the assumption that these counts come from a Poisson distribution, but we make the added assumption that the mean number of events, $\mu$, comes from a 2-parameter Gamma distribution.

\subsection{Mixture models}
Finite mixture (FM) modeling has emerged within the past 20 years as a popular means for handling unsupervised classification tasks.
The underlying principle behind mixture modeling is that we treat our observed data as having been sampled from a convex sum of distributions \cite{Bozdogan:1994, Titterington:1985}.
This concept is particularly useful for explainining overdispersion in count data as being due to an underlying heterogeneous population.

FM modeling has appeared in tandem with Poisson regression models.
The extent of this type of research has mostly been limited to zero-inflated, zero-truncated, and hurdle-type models \cite{Dohoo:2012}.
These models are designed to handle zero-counts, and are largely inappropriate for explaining heterogeneities due to covariates.
Papastamoulis \emph{et al.} (2014) described a methodology that can address covariate heterogeneity using FMs of Poisson regression models in the context of high-throughput sequencing data \cite{Papastamoulis:2014}.
Their EM algorithms for estimating $G > 10$ mixtures are freely available online from CRAN.

In addition, Park and Lord (2009) have developed FMs of Poisson and negative binomial regression models for explaining hetergeneities in vehicle crash data \cite{Park:2009}.
Of the modeling frameworks that have been suggested in the literature, theirs is most similar to ours in that they proposed using the information criteria AIC, BIC, and DIC to choose the optimal number of mixing components.
However, these criteria did not give conclusive results, and so they were forced to subjectively choose a mixture of 2 NB-2 models.
In the proceeding sections, we will discuss potential reasons for these inconclusive results.
For more information on finite mixtures of negative binomial regression models, see also Zou \emph{et al.} (2013) \cite{Zou:2013}.

\subsection{Model selection}
Model selection exists as an alternative approach to classical hypothesis-drive statistics which require distributional assumptions and an arbitrary selection of a confidence level to evaluate $p$-values.
Akaike was the first to propose a criterion for conducting model selection when he introduced his celebrated Akaike information criterion (AIC) in 1973.
The formula, which involves a tradeoff between a candidate model's lack of fit given by the maximized log-likelihood function, and a penalty term for the number of parameters, $n_k$, is shown below:
\begin{equation}
\textnormal{AIC} = -2\log L(\hat{\theta} | X) + 2n_k.
\end{equation}
When we compare 2 or more models that have been fit to the same set of data, we prefer to choose the model that minimizes the AIC score.

As it turns out, the original penalty term that Akaike proposed, $2n_k$, is not enough to prevent model overfitting \cite{Bozdogan:1987}.
Numerous other information criteria (IC) have been proposed, including but not limited to Schwarz's Bayesian Criterion (SBC or BIC) \cite{Schwarz:1978}, the deviance information criterion (DIC) (which we will not consider here), the consistent form of AIC (CAIC) \cite{Bozdogan:1987}, and the information-theoretic measure of complexity (ICOMP) \cite{Bozdogan:1994} which involves evaluating the inverse Fisher's information matrix (IFIM).
The reader is directed to the respective citations for derivations of the formulae.
We will point out that each of the criteria we consider in this paper all share the lack-of-fit term in common -- what makes them different is how they penalize overparameterization.

\section{Methods}

\subsection{Log-likelihood functions}
Our ultimate goal when we carry out model selection in this framework is to determine the optimal number of components in a FM model, given a fixed maximum number of components (see \cite{Bozdogan:1994} for heuristics on choosing this number).
This optimal number is reflective of the number of subpopulations composing the entire population from which we have sampled our data.
Before we can perform model selection, we must be able to generate the various criteria discussed previously.
To do so requires that we compute the log-likelihood function.

Recall equation (\ref{poissmodel}) when dealing with Poissonian data whose mean and variance are similar.
It follows that $\mathbb{E}[Y | X, \beta] = \exp(\beta^T X)$.
Since our outcome is sampled from a convex sum of Poisson distributions we have the following probability for observing $n$-dimensional count data, $Y$:

\begin{equation}\label{poisspdf}
\mathbb{P}[Y; X, \beta, \pi] = \prod_{i = 1}^n \sum_{g = 1}^G \pi_g \frac{\exp(y_i \beta^T_g x_i) \exp\left(-\exp(\beta^T_g x_i) \right)}{y!}.
\end{equation} 

\noindent In equation (\ref{poisspdf}), $G$ is the total number of components in the FM model; $\pi_g$ is the weight assigned to component $g$; and, $\beta_g$ refers to the effects in the $g$-th component, as this will vary within components.
It follows directly from equation (\ref{poissmodel}) that the log-likelihood is
\begin{equation}\label{LLpoiss}
\log\left(L(\beta, \pi; Y, X)\right) = \displaystyle \sum_{i = 1}^n \left\{ \sum_{g = 1}^G \left[ \log(\pi_g) + y_i \beta^T_g x_i + \exp(-\beta_g^T x_i) - \log(y_i!) \right] \right\}.
\end{equation}

When data are significantly overdispersed, recall that we assume our $n$ counts, $Y$, come from a Poisson distribution whose mean parameter, $\lambda$, is Gamma-distributed.
This additional constraint gives us the following probability density for $Y$:

\begin{equation}\label{nbpdf}
\mathbb{P}[Y; X, \beta, \pi, \alpha] = \displaystyle \prod_{i=1}^n \sum_{g = 1}^G \left\{ \pi_g \frac{\Gamma\left(y_i + 1 / \alpha_g \right)}{\Gamma\left(1 / \alpha_g\right) y_i!} 
\left( \frac{\alpha_g \beta_g^T x_i}{1 + \alpha_g \beta^T_g x_i}\right)^{y_i}
 \left(\frac{1}{1 + \alpha_g \beta_g^T x_i}\right)^{\frac{1}{\alpha_g}} \right\}.
\end{equation}

\noindent The interpretation of $\pi$ and $\beta$ follows as before.
The $\alpha_i$ are overdispersion parameters, satisfying $\textnormal{var}(Y) = (1 + \alpha \beta^T X) \beta^T X$ (note: $\textnormal{mean}(Y) = \beta^T X$).
The quadratic dependence of the variance on the mean is why we refer to these as NB-2 models.

Finally, following directly from equation (\ref{nbpdf}), the log-likelihood function for the negative binomial regression model is shown below \cite{Biswas:2013}:
\begin{equation}\label{LLnb}
\def\arraystretch{1.4}
\begin{array}{rcl}
\log\left(L(\beta, \pi, \alpha; Y, X)\right) & = &  \displaystyle \sum_{i = 1}^n \left\{ \sum_{g = 1}^G \left[ \log\left(\Gamma\left(y + \frac{1}{\alpha_g} \right)\right) - \log \left(\Gamma\left(\frac{1}{\alpha_g}\right)\right) +  y_i \left( \log(\alpha_g \beta_g^T x_i) \right. \right. \right. \\[3ex]
& & \left. \left. \left. - \log(1 + \alpha_g \beta_g^T x_i)\right) - \left(\frac{1}{\alpha_g}\right) \log(1 + \alpha_g \beta_g^T x_i) - \log(y_i!) \right] \right\}.
\end{array}
\end{equation}

Note that when $\alpha = 0$, the data are not overdispersed and so the assumption that $\textnormal{mean}(Y) = \textnormal{var}(Y)$ is satisfied.
Using our methodology, if the overdispersion parameter is below a certain threshold, we elect to perform Poisson regression, otherwise we perform NB-2 regression.
This is especially important for computing the log-likelihood functions once we carry out scoring.

\subsection{Initialization}
The choice of an appropriate initialization scheme was a source of great concern for Papastamoulis \emph{et al.} in their algorithm \cite{Papastamoulis:2014}.
They used a ``Small-EM'' strategy from Biernacki \emph{et al.} (2003), which was necessary for dealing with a large number of component mixtures ($G >> 5$) \cite{Biernacki:2003}.

We are not interested in exploring the possibility of more than $G = 5$ components unless there is some justifiable epidemiological reason for doing so.
Because of this, we opted for using a Poisson mixture model to perform unsupervised classification of our observed counts (see \cite{Massaro:2015a}).
As a result, our methodology involves first sorting the count data into $G \leq 5$ groups, and then assigning observations in each group to an initial Poisson or NB-2 regression model.

\section{Results}
All numerical simulations were carried out in MATLAB 2015a.
We used ArcGIS 10.2 for generating the map of Tennessee counties.

\subsection{Simulated data}
We begin using our proposed model framework to choose the optimal number of components in a simple set of simulated data.
Each observation in the data is an ordered pair, $(x, y)$, where $x \sim N(\mu_i, 1)$, and $y \sim \textnormal{Poisson}(\lambda_i)$, $i = 1, \dots, 4$.
The parameters $\mu_i$ and $\lambda_i$ for all $i$ are randomly generated integers between 1 and 50. 
In this way, each observation comes from 1 of 4 potential groups.

\begin{figure*}[!ht]
\centering
\begin{tabular}{cc}
\includegraphics[width=2.5in]{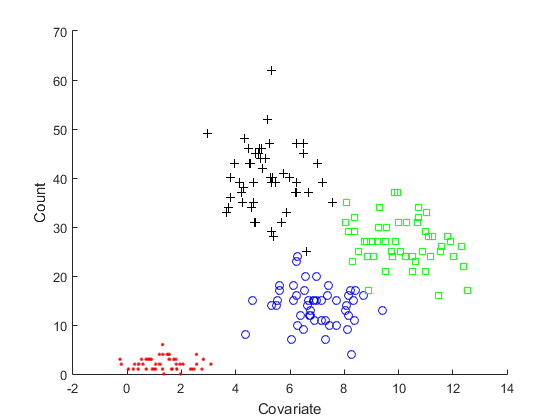}
\label{fig_first_case} & \includegraphics[width=2.5in]{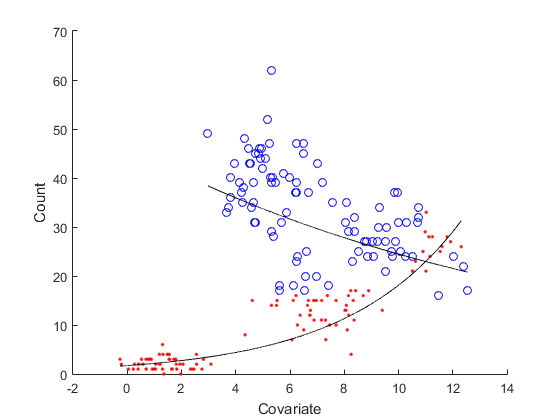}
\label{fig_second_case}
\end{tabular}
\caption{On the left, we see the data in its original form, with each observations coming from 1 of 4 distinct groups.  On the right, we see the same data, with the mixture of two Poisson regression models that was chosen by ICOMP overlaid.  Different colors and symbols indicate observations belonging to different mixtures.}
\label{fig:sim}
\end{figure*}

We tried fitting FMs of 1 to 5 Poisson regression models to these data; their scores are shown in Table \ref{tab:sim}.
We see that all 3 scores select $G = 2$ as the optimal number of components.
The regression models corresponding to the solution when $G = 2$ are shown in the right pane of Figure \ref{fig:sim}.
The regression models each seem to do well modeling these data.

\begin{table*}[!ht]
\centering
\def\arraystretch{1.4}\itemsep=1.4pt
\begin{tabular}{c|ccc} \hline \hline
$G$ & AIC & CAIC & ICOMP \\ \hline
1 & 2931.14 & 2939.74 & 2944.77 \\
2 & 1392.98 & 1414.47 & 1416.78 \\
3 & 1400.08 & 1434.46 & 1434.19 \\
4 & 1428.51 & 1475.79 & 1472.49 \\
5 & 1455.42 & 1515.59 & -- \\
\hline \hline
\end{tabular}
\caption{IC scores for mixtures of $G = 1, \dots, 5$ Poisson regression models applied to the simulated data shown in Figure \ref{fig:sim}.  All three of the scores are minimized for $G = 2$.  Our regression framework failed to produce a stable IFIM when $G = 5$, so no ICOMP score was computed.}
\label{tab:sim}
\end{table*}

\subsection{Tennessee HIV counts}
We now use real data taken from a 2014 county-level study on persons living with HIV in the U.S. South \cite{Gray:2015}.
Readers are encouraged to refer to \cite{Gray:2015} for a summary of the entire data set.
For demonstrating our methodology, we are interested only in the TN counties that were included in the study.
These data are summarized in Table \ref{tab:TN}.
For reference, the abbreviations we use when referring to individual covariates are summarized below in Table \ref{tab:abbr}.

\begin{table}[!ht]
\centering
\def\arraystretch{1.4}
\begin{tabular}{cllc}
\hline \hline
No. & ID & Description & Data type \\ \hline
-- & HIV & Number of persons living with HIV & Continuous (count) \\
1 & SCH & Proportion of persons with less than a HS education & Continuous \\
2 & POV & Proportion of persons living below the poverty line & Continuous \\
3 & INC & Natural logarithm of median income & Continuous \\
4 & URB & Urbanicity indicator (population $<$50,000/$>$50,000) & Categorical (2) \\
5 & UMP & Unemployment rate & Continuous \\
6& NHB & Proportion of Non-Hispanic Black persons & Continuous \\
7 & NHW & Proportion of Non-Hispanic White persons & Continuous \\
8 & HSP & Proportion of Hispanic persons & Continuous \\
\hline \hline
\end{tabular}
\caption{Description of the variables in the Tennessee HIV data.  The outcome is HIV.  Categorical variables have the number of levels listed in parentheses.}
\label{tab:abbr}
\end{table}

\begin{figure*}[!ht]
\centering
\includegraphics[width=3.5in]{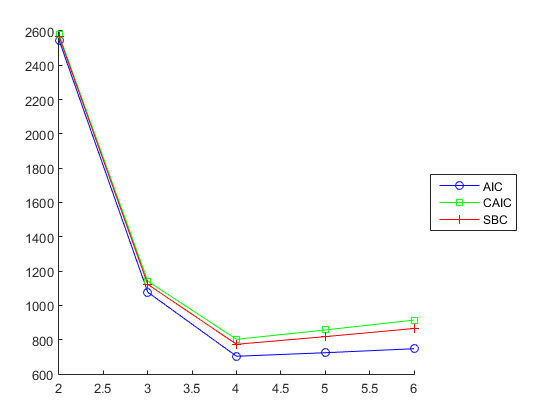}
\caption{Information criteria scores for finite mixtures of $G = 1, \dots, 5$ NB-2 models using the Tennessee county-level HIV data.  We see that all 3 scores are minimized for $G = 3$.}
\label{fig:TN}
\end{figure*}

The count data are largely overdispersed (mean(HIV) = 166.45, var(HIV) $>$ 500,000), so we proceed by using mixtures of NB-2 models.
In Figure \ref{fig:TN}, we see that the model selected 4 regression models, which is indicative of 4 separate homogeneous subpopulations within Tennessee counties.

Given that there are 4 subpopulations, we used a genetic algorithm (GA) to select the subset of variables from the original 8 that contributes most to sorting the data.
GA have been used numerous times for carrying out variable subset selection, although never for Poisson or NB-2 regression (see \cite{Akbilgic:2011, Broadhurst:1996, Vinterbo:1999} for GA applied to other regression models).
The results for performing GA subsetting in our model are summarized in Table \ref{tab:GA}.
We see that variables 4 and 7 appear in each of the subsets most frequently chosen, and the model fitted with only variables 1, 3, 4, 5, and 7 is the best at minimizing the IC scores.

\begin{table}[!ht]
\centering
\def\arraystretch{1.4}
\begin{tabular}{lrccc}
\hline \hline
Subset & Rel. freq. & AIC & CAIC & SBC \\ \hline
1, 3, 4, 5, 7 & 17.00\% & 708.76 & 776.89 & 756.89 \\
1, 2, 3, 4, 5, 7 & 10.00\% & 703.51 & 781.86 & 758.86 \\
4, 7 & 10.50\% & 849.86 & 887.33 & 876.33 \\ 
\hline \hline
\end{tabular}
\caption{Relative frequencies and IC scores for variable subsets chosen using the GA.}
\label{tab:GA}
\end{table}

In the interest of parsimony, Table \ref{GA_coef} shows coefficients from each of the 4 components in a FM model fitted with just the urban indicator and the non-Hispanic white proportion.
We see that the proportion of whites in a county is significantly protective against HIV count in 3 out of 4 components, while the urban indicator is a significant risk for HIV count.

In components 1 and 4, urban indicator was not included in the regression analysis because all of the counties belonging to those components had urban indicators of 0 or 1, respectively.
Based on Table \ref{summ_comp}, this makes sense.
We see that component 1 contains counties with HIV counts less than 15, while component 4 contains counties with counts greater than 100.
The proportion of individuals with a high school diploma or lower decreases in each component, as do the non-Hispanic white proportions.
Interestingly, the poverty rate and unemployment rates also decrease.
At the same time, the urban indicator, and the non-Hispanic black and Hispanic proportions increase with each component.

\section{Discussion}
Our results indicate that the proportion of white persons in a county seems to be a significant predictor of HIV count in that county.
This is slightly inconsistent with the conclusions of Gray, \emph{et al.} \cite{Gray:2015}, who found that the proportion of black persons was the most significant contributor to HIV rate.
That being said, our GA the non-Hispanic black variable was selected numerous times as a significant variable, however it was not one the most significant (results not shown).

\begin{figure*}[!ht]
\centering
\includegraphics[width=6in]{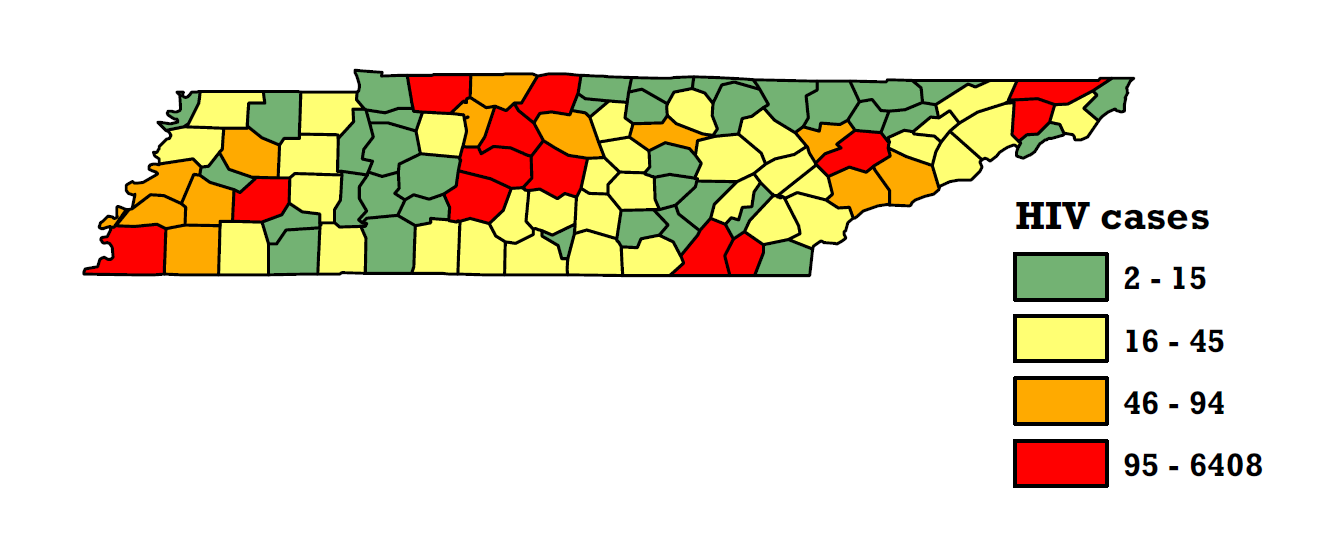}
\caption{HIV cases by county in Tennessee.}
\label{fig:map}
\end{figure*}

That urban indicator was selected as a significant variable comes as no surprise.
This result tells us that as the population increases, we should also expect to see an increase in HIV count.
For numerous different reasons, this result makes perfect sense.
Moreover, in Figure \ref{fig:map}, we see that the counties with the highest HIV counts correspond to the some of the largest cities in Tennessee, including Knox (Knoxville), Davidson (Nashville), Hamilton (Chattanooga), and Shelby (Memphis).

It was also interesting to find that poverty rates and education levels tended to be negatively associated with HIV counts.
Our best explanation is that, at least in Tennessee, population size is a confounder for education level and poverty rate.
That is to say that as the population of a county increases, we also expect to see access to jobs and education increasing.
It would be worth investigating these results further to determine what potential public health impacts exist.

\section{Conclusion}
We have proposed a novel framework for analyzing data sets in which the response variable is a count.
Our approach allows us to systematically identify homogeneous subpopulations arising in the dataset, based on the covariates and their contributions to the count data in a Poisson regression model.
We are also able to select the covariates that contribute most to determining the aforementioned subpopulations.
Using this approach on a set of HIV count data in Tennessee has led to reasonable and quantifiable results.
This would suggest that we could apply this approach to other sets of data involving counts of other infectious diseases, such as influenza, dengue fever, or even the Ebola virus disease.


\newpage
\section{Figures}

\newpage
\section{Tables}
\begin{table*}[!ht]
\centering
\def\arraystretch{1.4}
\begin{tabular}{lcccc} \hline \hline
Name & Mean & St. Dev. & Min & Max \\ \hline
HIV&    166.45&    770.43&      2.00&   6408.00
\\
SCH&   0.21&      0.05&      0.05&      0.30
\\
POV&     0.17&      0.04&      0.05&      0.30
\\
INC&  10.55&      0.20&     10.01&     11.42
\\
URB&   0.31&      0.46&      0.00&      1.00
\\
UMP&   0.11&      0.03&      0.05&      0.18
\\
NHB&   0.07&      0.10&      0.00&      0.52
\\
NHW&   0.88&      0.12&      0.39&      0.99
\\
HSP&   0.03&      0.02&      0.00&      0.11
\\
\hline\hline
\end{tabular}
\caption{Summary statistics of Tennessee county-level HIV data taken from \cite{Gray:2015}.}
\label{tab:TN}
\end{table*}

\begin{table*}[!ht]
\centering
\def\arraystretch{1.4}
\begin{tabular}{lcccc} \hline \hline
Variable & $\beta$ & S. E. & 2.5\% & 97.5\% \\ \hline \hline
\multicolumn{5}{c}{Component 1} \\ \hline
URB & -- & -- & -- & -- \\
NHW & -2.14 & 1.30 & -4.69 & 0.41 \\
Int & 4.06 & 1.20 & 1.70 & 6.41 \\
& & & & \\
\multicolumn{5}{c}{Component 2} \\ \hline
URB & 0.40 & 0.091 & 0.22 & 0.58 \\
NHW & -1.85 & 0.51 & -2.85 & -0.85 \\
Int & 4.84 & 0.45 & 3.96 & 5.72 \\
& & & & \\
\multicolumn{5}{c}{Component 3} \\ \hline
URB & 0.35 & 0.11 & 0.12 & 0.57 \\
NHW & -1.05 & 0.39 & -1.81 & -0.29 \\
Int & 4.94 & 0.28 & 4.38 & 5.50 \\
& & & & \\
\multicolumn{5}{c}{Component 4} \\ \hline
URB & -- & -- & -- & -- \\
NHW & -7.12 & 1.40 & -9.86 & -4.37 \\
Int & 11.57 & 1.08 & 9.46 & 13.68 \\
\hline \hline
\end{tabular}
\caption{Coefficients from the Poisson regression models in each component, using the variables selected by GA.}
\label{GA_coef}
\end{table*}

\newpage
\begin{table*}[!Ht]
\centering
\footnotesize
\def\arraystretch{1.4}
\begin{tabular}{lcccc} \hline \hline
Variable & Mean & St. dev. & Min. & Max. \\ \hline \hline
\multicolumn{5}{c}{Component 1} \\ \hline
HIV &      8.03&      4.00&      2.00&     15.00
\\
SCH &      0.24&      0.04&      0.16&      0.30
\\
POV &   0.20&      0.04&      0.10&      0.30
\\
INC &     10.43&      0.15&     10.01&     10.71
\\
URB &      0.00&      0.00&      0.00&      0.00
\\
UMP &     0.12&      0.03&      0.07&      0.17
\\
NHB &      0.03&      0.05&      0.00&      0.26
\\
NHW &      0.93&      0.06&      0.68&      0.99
\\
HSP &     0.02&      0.02&      0.00&      0.09
\\
& & & & \\
\multicolumn{5}{c}{Component 2} \\ \hline
HIV&     27.55&      8.89&     15.00&     45.00
\\
 SCH&     0.21&      0.03&      0.15&      0.30
\\
POV&      0.17&      0.03&      0.12&      0.23
\\
INC&     10.54&      0.10&     10.30&     10.81
\\
URB&      0.27&      0.45&      0.00&      1.00
\\
UMP&      0.11&      0.02&      0.08&      0.18
\\
NHB&      0.05&      0.07&      0.00&      0.41
\\
NHW&      0.89&      0.08&      0.56&      0.97
\\
HSP&      0.03&      0.03&      0.01&      0.11
\\
& & & & \\
\multicolumn{5}{c}{Component 3} \\ \hline
HIV&     72.92&     16.84&     45.00&     94.00
\\
SCH&      0.17&      0.04&      0.12&      0.26
\\
POV&   0.14&      0.05&      0.09&      0.23
\\
INC& 10.70&      0.21&     10.40&     11.02
\\
URB&  0.54&      0.52&      0.00&      1.00
\\
UMP&  0.10&      0.02&      0.07&      0.15
\\
NHB&  0.14&      0.15&      0.01&      0.49
\\
NHW&  0.81&      0.15&      0.45&      0.94
\\
HSP&  0.03&      0.01&      0.02&      0.06
\\
& & & & \\
\multicolumn{5}{c}{Component 4} \\ \hline
HIV&   1051.31&   1913.88&    101.00&   6408.00
\\
SCH&  0.13&      0.03&      0.05&      0.18
\\
POV&  0.13&      0.03&      0.05&      0.17
\\
INC& 10.80&      0.21&     10.60&     11.42
\\
URB&  1.00&      0.00&      1.00&      1.00
\\
UMP&  0.09&      0.02&      0.05&      0.12
\\
NHB&  0.16&      0.15&      0.02&      0.52
\\
NHW&  0.75&      0.16&      0.39&      0.94
\\
HSP&  0.05&      0.02&      0.01&      0.10
\\
\hline\hline
\end{tabular}
\caption{Summary statistics for counties belonging to each component.}
\label{summ_comp}
\end{table*}

\end{document}